\documentclass[sn-mathphys]{sn-jnl}

\jyear{2021}%

\theoremstyle{thmstyleone}%
\theoremstyle{thmstyletwo}%
\newcommand{\TT}{\textcolor{black}}
\theoremstyle{thmstylethree}%

\begin{document}

\title[Article Title]{OceanNet: A principled neural operator-based digital twin for regional oceans}


\author*[1]{\fnm{Ashesh} \sur{Chattopadhyay}}\email{aschatto@ucsc.edu}

\author[2]{\fnm{Michael} \sur{Gray}}\email{magray@ncsu.edu}
\author[2]{\fnm{Tianning} \sur{Wu}}\email{twu27@ncsu.edu}
\author[2]{\fnm{Anna B.} \sur{Lowe}}\email{ablowe@ncsu.edu}
\author*[2]{\fnm{Ruoying} \sur{He}}\email{rhe@ncsu.edu}


\affil*[1]{\orgdiv{Applied Mathematics}, \orgname{University of California, Santa Cruz}, \orgaddress{\city{Santa Cruz}, \postcode{95060}, \state{California}, \country{United States}}}

\affil[2]{\orgdiv{Marine, Earth \& Atmospheric Sciences}, \orgname{North Carolina State University}, \orgaddress{ \city{Raleigh}, \postcode{27695}, \state{North Carolina}, \country{United States}}}

\abstract{While data-driven approaches demonstrate great potential in atmospheric modeling and weather forecasting, ocean modeling poses distinct challenges due to complex bathymetry, land, vertical structure, and flow non-linearity. This study introduces OceanNet, a principled neural operator-based digital twin for \TT{regional sea-suface height emulation}. OceanNet uses a Fourier neural operator and predictor-evaluate-corrector integration scheme to mitigate autoregressive error growth and enhance stability over extended time scales. A spectral regularizer counteracts spectral bias at smaller scales. OceanNet is applied to the northwest Atlantic Ocean western boundary current (the Gulf Stream), focusing on the task of seasonal prediction for Loop Current eddies and the Gulf Stream meander. Trained using historical sea surface height (SSH) data, OceanNet demonstrates competitive forecast skill \TT{compared to a} state-of-the-art dynamical ocean model forecast, reducing computation by 500,000 times. These accomplishments demonstrate \TT{initial steps for} physics-inspired deep neural operators as cost-effective alternatives to high-resolution numerical ocean models.
}

\maketitle
\section{Introduction}\label{sec1}

An essential component of the Earth’s ocean circulation is the northern Atlantic Ocean’s western boundary current (WBC) system, which includes the Loop Current (LC), Loop Current eddies (LCE), the Gulf Stream (GS), and Gulf Stream meander (GSM). The WBC plays a critical role in the transport of heat, salt, and nutrients, and strongly influences global weather, climate, and marine ecosystems. Recent studies have shown the promise of using data-driven approaches to model and predict atmospheric circulation, but complexities in the ocean system present additional challenges when building data-driven models~\cite{pathak2022fourcastnet, lam2022graphcast, bi2022pangu}. One challenge is the complex land boundaries, such as in the Gulf of Mexico (GoM). Equally difficult to model, the GS frequently sheds eddies as a result of nonlinear flow and the interactions of cool, subpolar circulation from the north and warm, subtropical circulation from the south.  Predicting LCEs in the GoM and GS meander and eddy separation have been long-standing challenges for numerical ocean models~\cite{dengo1993problem,chassignet2008gulf,ezer2016revisiting}. The separation mechanism involves nonlinear barotropic and baroclinic instabilities of ocean currents, complex flow-topography interactions, and, in the case of the GS, interaction with the Deep Western Boundary Current~\cite{spall1996dynamics,zhang2007role,hurlburt2008gulf,schoonover2017local}. These challenging features require carefully calibrated, high-resolution numerical models and tremendous computing resources to resolve~\cite{andres2016recent,gangopadhyay2019observed,silver2023increased}. 

Recent efforts in emulating ocean dynamics with deep learning-based
approaches have primarily focused on predicting large-scale circulation features, such as those resolved by  empirical orthogonal functions (EOFs)~\cite{wang2019medium} or on constructing low-dimensional representations~\cite{agarwal2021comparison}. However, there has yet to be a comprehensive data-driven model for the ocean akin to those used in global atmospheric modeling. 

 Data-driven modeling in Earth sciences has focused on predicting global atmospheric dynamics~\cite{chattopadhyay2019analog,chattopadhyay2020deep,pathak2022fourcastnet,lam2022graphcast,bi2022pangu}; weather prediction has been used as a benchmark. The primary advantage of data-driven models lies in their computational efficiency, enabling the execution of a large number of ensembles. This facilitates seamless and efficient data assimilation leading to improved estimates of extreme weather events. Several recent studies~\cite{pathak2022fourcastnet,lam2022graphcast,bi2022pangu} have demonstrated that these data-driven models can achieve accuracy comparable to, and often surpassing, state-of-the-art numerical weather prediction models such as the Integrated Forecasting System (IFS), at a significantly reduced computational cost. This achievement holds substantial promise for incorporating deep learning-based data-driven models into the suite of operational weather forecasting models. 

A significant limitation of these models arises when they are integrated over longer time scales ($\geq$ 2 weeks), leading to instability and the emergence of nonphysical features (see Fig. 1 in Chattopadhyay et al.~\cite{chattopadhyay2023long}). The cause of this instability was identified as “spectral bias”~\cite{chattopadhyay2023long}, an inductive bias in \textit{all} deep neural networks that hinders their ability to capture small-scale features in turbulent flows. A potential solution was proposed in~\cite{chattopadhyay2023long} in the form of a framework to construct long-term stable digital twins for atmospheric dynamics.

Research into data-driven digital twins for both global and regional ocean modeling is still in its infancy and has not been explored to the same extent as in atmospheric modeling. The development of long-term stable, accurate data-driven ocean models that maintain computational efficiency is essential for realizing data-driven digital twins for the entire Earth system. 

We present here a principled neural operator-based digital twin \TT{to predict the sea-surface height (SSH) of} the northwest Atlantic Ocean’s western boundary current, built upon the same principles as the FouRK model introduced in Chattopadhyay \textit{et al.}~\cite{chattopadhyay2023long}. The proposed model, OceanNet, relies on a Fourier neural operator (FNO), which incorporates a predictor-evaluate-corrector (PEC) integration scheme to suppress autoregressive error growth. Additionally, a spectral regularizer is employed to mitigate spectral bias at small scales. 

OceanNet is trained on historical SSH data from a high-resolution northwest Atlantic Ocean reanalysis and demonstrates \TT{long-term} stability and competitive skills. It \TT{remains competitive with} SSH prediction made by a regional ocean dynamical prediction that is based on the state-of-the-art Regional Ocean Modeling System (ROMS) while maintaining a computational cost that is 500,000 cheaper. \TT{While OceanNet only emulates high-resolution SSH fields for the regional ocean, future work incorporating other key variables in ocean dynamics would underscores the potential of structured, physics-inspired deep neural operators as a cost-effective alternative to computationally expensive high-resolution numerical models.}

\begin{figure}[h]%
\centering
\includegraphics[width=0.9\textwidth]{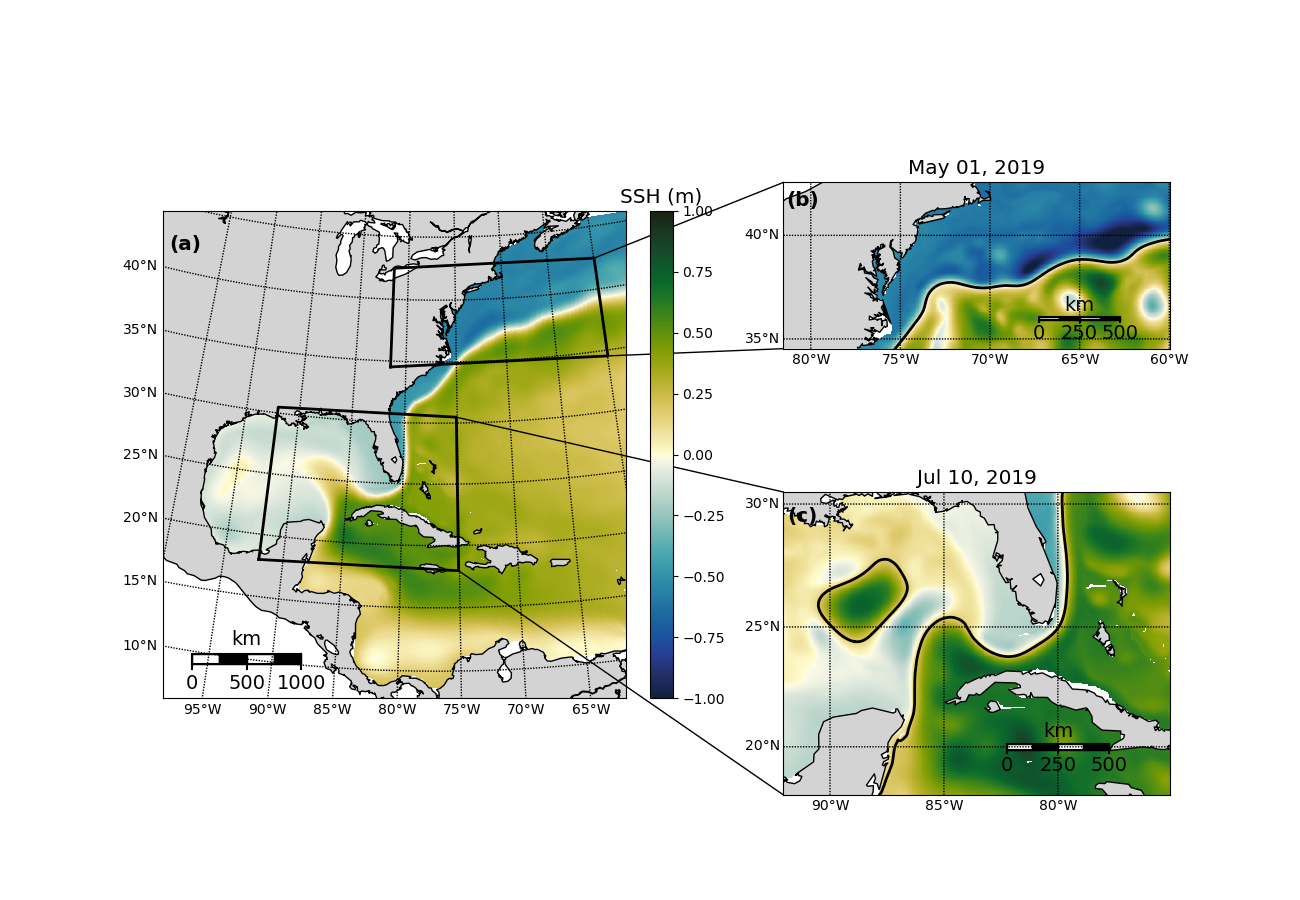}
\caption{ The domain for the reanalysis data covering the northwestern Atlantic (a). The two subdomains used to develop OceanNet, specifically (b) the GS separation from the central US east coast to 60$^{\circ}$W and (c) the  LC eddy-shedding region in the eastern GoM, extending from 92$^{\circ}$W into the Atlantic 75$^{\circ}$W. The mean SSH from from 1993-2020 in the reanalysis data is shown in (a), while (b) and (c) depict daily mean SSH on May 1, 2019, and July 10, 2019, respectively. All three domains share the same color scale.}
\label{fig_domains}
\end{figure}

\section{Results}\label{sec2}
Mesoscale ocean circulation dynamics can be well represented by the spatio-temporal evolution of SSH fields. In this section we conduct a comprehensive comparison of SSH predictions generated by OceanNet and the regional ocean dynamical model forecast with independent reanalysis data. To assess the performance rigorously, we employ both qualitative and quantitative measures. The metrics for evaluating predictive accuracy of SSH include root mean squared error (RMSE) and correlation coefficient (CC), which are widely recognized and employed in forecasting~\cite{pathak2022fourcastnet,bi2022pangu,lam2022graphcast,chattopadhyay2023long}. In addition, we incorporate a specialized object-tracking metric to evaluate the prediction of major ocean features delineated by SSH contours: the modified Hausdorff distance (MHD, \cite{dukhovskoy2015skill}). MHD quantifies the comparison of predicted eddy/meander structures to their counterparts in the reanalysis data; identical \textit{shapes} \TT{at the same time instant} have an MHD of zero. To provide a comprehensive assessment, we also present qualitative snapshots of the predicted SSH fields generated by OceanNet, regional ocean dynamical forecast, and the independent SSH field derived from the reanalysis. 

\subsection{Performance of OceanNet in the GoM}

\begin{sidewaysfigure}[htbp]%
\centering
\includegraphics[width=0.9\textwidth]{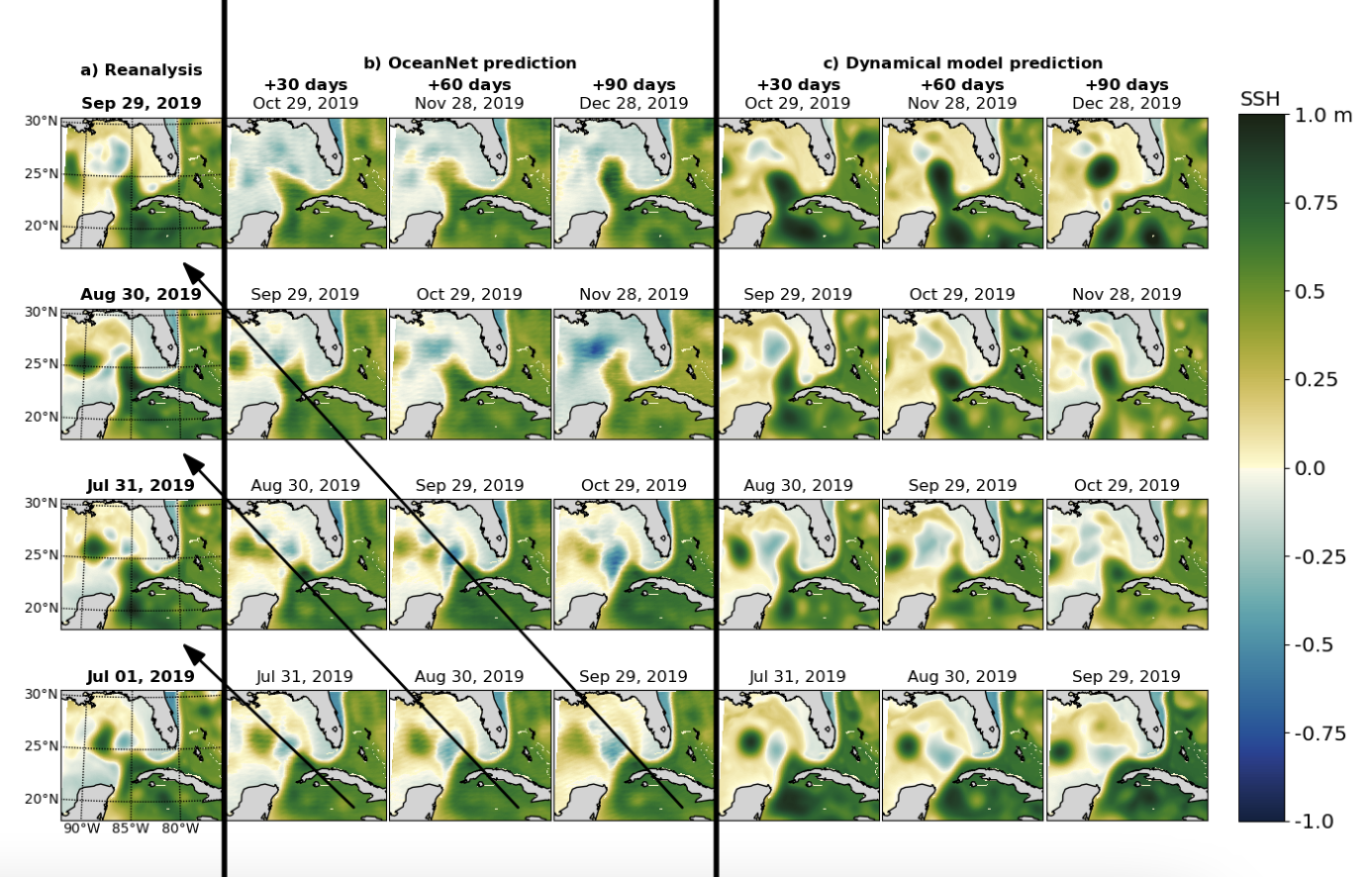}
\caption{Prediction performance of OceanNet. (a) SSH fields from the ocean reanalysis. (b) Predicted SSH generated by the OceanNet. (c) regional ocean dynamical model forecasts. Each row in the OceanNet and dynamical model predictions was initialized with the corresponding reanalysis data in the left column.  The predictions display SSH forecasts for 30, 60, and 90 days.  To evaluate the predictions, we can examine a diagonal comparison with the reanalysis SSH, as indicated by the black arrows in (b). The same diagonal comparison can be conducted with the ocean reanalysis data for (c).}
\label{fig:fig_diagGOM}
\end{sidewaysfigure}

We present an evaluation of OceanNet’s seasonal (up to 120 days) forecasting performance compared to that of a regional ocean dynamical model forecast for two major GoM LCE shedding events: Eddy Sverdrup (Jul 2019-Jan 2020) and Eddy Thor (Jan 2020 - Sep 2020). The target LC dynamics, represented by spatio-temporal evolutions of SSH, are from the reanalysis dataset. Specifics regarding the generation and structure of the reanalysis dataset are covered in section~\ref{sec:ocean_reanalysis}. 

A qualitative assessment (Fig.~\ref{fig:fig_diagGOM}) reveals that OceanNet \TT{performs relatively well in terms of maintaining the general physical consistency of the emulation of the ocean dynamics. However, there are certain cases, where ROMS does signficantly better than OceanNet. For example, the eddy configuration in September $29$, $2019$, OceanNet predictions of the eddy diffuses quickly while ROMs maintains the strength of the eddies. However, e.g., the eddy cofiguration of Aug $30$, $2019$, is better captured by OceanNet, while ROMs predicts an unphysical low-pressure region just off the coast. Generally for the GoM region, both RMSE and CC for OceanNet and ROMs is roughly the same, while the eddy shape is better captured by OceanNet as revealed by the MHD.} OceanNet, however, maintains both stability and physical consistency for at least 120 days. For this domain, OceanNet is trained with a $5$-day lead, consistent with the LC and LCE evolution timescale. ROMS based regional dynamical model predictions were initialized five days apart to allow for fair comparisons between the two models. OceanNet predictions have a ``wall-clock" execution time of microseconds, approximately 500,000 times faster than regional dynamical model forecasts. \TT{So, while ROMS, in this case, is just as accurate or even better for some of the eddy shedding cases, the computational advantage of OceanNet is undeniable.}

\begin{figure}
\noindent\includegraphics[width=\textwidth]{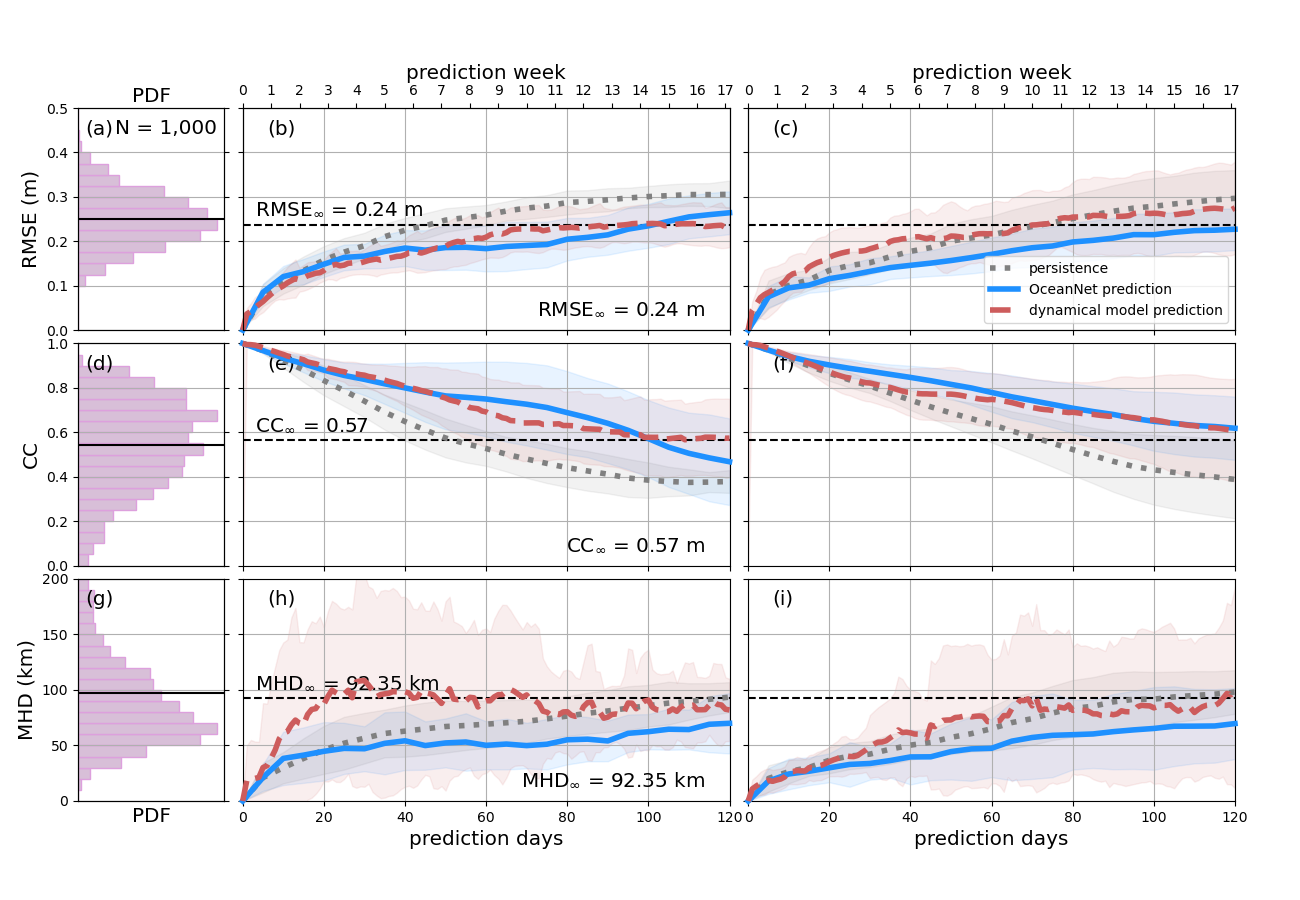}
\caption{OceanNet performance metrics in the GoM: RMSE (a-c), CC (d-f), and MHD (g-i). In the left column, probability density functions (PDFs) are presented for 1,000 random pairs sourced from the training data spanning 1993-2018 (means shown by black horizontal lines). The performance statistics, calculated based on forecasts of 0-120 days for eddies Sverdrup (middle column) and Thor (right column), are displayed as mean values (lines) with standard deviations (shading). The black horizontal dashed lines in these columns denote saturation values determined as 95\% of the means derived from the random pairs. In the middle and right columns, solid blue lines are OceanNet, dashed red lines are the regional ocean dynamical forecast model, and gray dots are persistence. These representations illustrate how each method’s statistics compare with the target SSH from the reanalysis dataset. \TT{The shading is obtained from the standard deviation due to the choice of different initial conditions to evluate the forecast skill.}}
\label{fig:RMSE_CC_MHD}
\end{figure}

An examination of RMSE, CC, and MHD for SSH, over a 120-day period (Fig.~\ref{fig:RMSE_CC_MHD}) demonstrates that OceanNet is competitive \TT{with} regional dynamical model forecasts, and outperforms random predictions (indicated by the horizontal dashed lines). These random predictions are derived from 1,000 random pairs of samples from the training data covering the period of 1993-2018 to calculate the saturation values, which quantify the intrinsic predictability of the ocean system \cite{dalcher1987error, delsole2004predictability, dukhovskoy2023assessment}. \TT{RMSE values above random predictions indicate that the model has lost all short-term forecasting skill, as is obvious for chaotic dynamical systems. Beyond $100$ days, ROMs clearly show an improvement in the RMSE (CC) values as compared to OceanNet, however those values are higher (lower) than the saturation values and hence do not provide any insights into predictive performance.} Comparing OceanNet and regional dynamical model with persistence forecasting, we see that regional dynamical model suffers from inaccuracies in predicting the timing and configuration of the eddy-shedding and reattachment process, and thus underperforms persistence forecast on several occasions. 

We conducted a further comparison (Supplementary Information) of OceanNet with previous deep learning-based models that have been successfully used in weather prediction, such as the model introduced by Weyn et al.~\cite{weyn2020improving}.  The spectral characteristics of these models exhibit deviations in the high wavenumbers, rendering their predictions nonphysical beyond a $2$-week forecasting horizon (Fig. S1). A geostrophic constraint within the loss function was also tested (Fig. S2), but OceanNet in its current design performed better.

\subsection{Performance of OceanNet in the North Atlantic GS Region}

We present qualitative and quantitative evaluations of OceanNet’s forecasting performance compared to the regional ocean dynamical model forecast for GSM in the northwest Atlantic. OceanNet was specifically trained for this domain with a lead time of $4$ days to resolve the evolution of the northern boundary of the GS. To ensure alignment with the initialization dates for regional ocean model (which were separated by $5$ days), we show the $20$th, $40$th, and $60$th-day forecasts, as opposed to the $30$th, $60$th, and $90$th-day forecasts used for the LC. 

\begin{sidewaysfigure}[htbp]%
\centering
\includegraphics[width=0.9\textwidth]{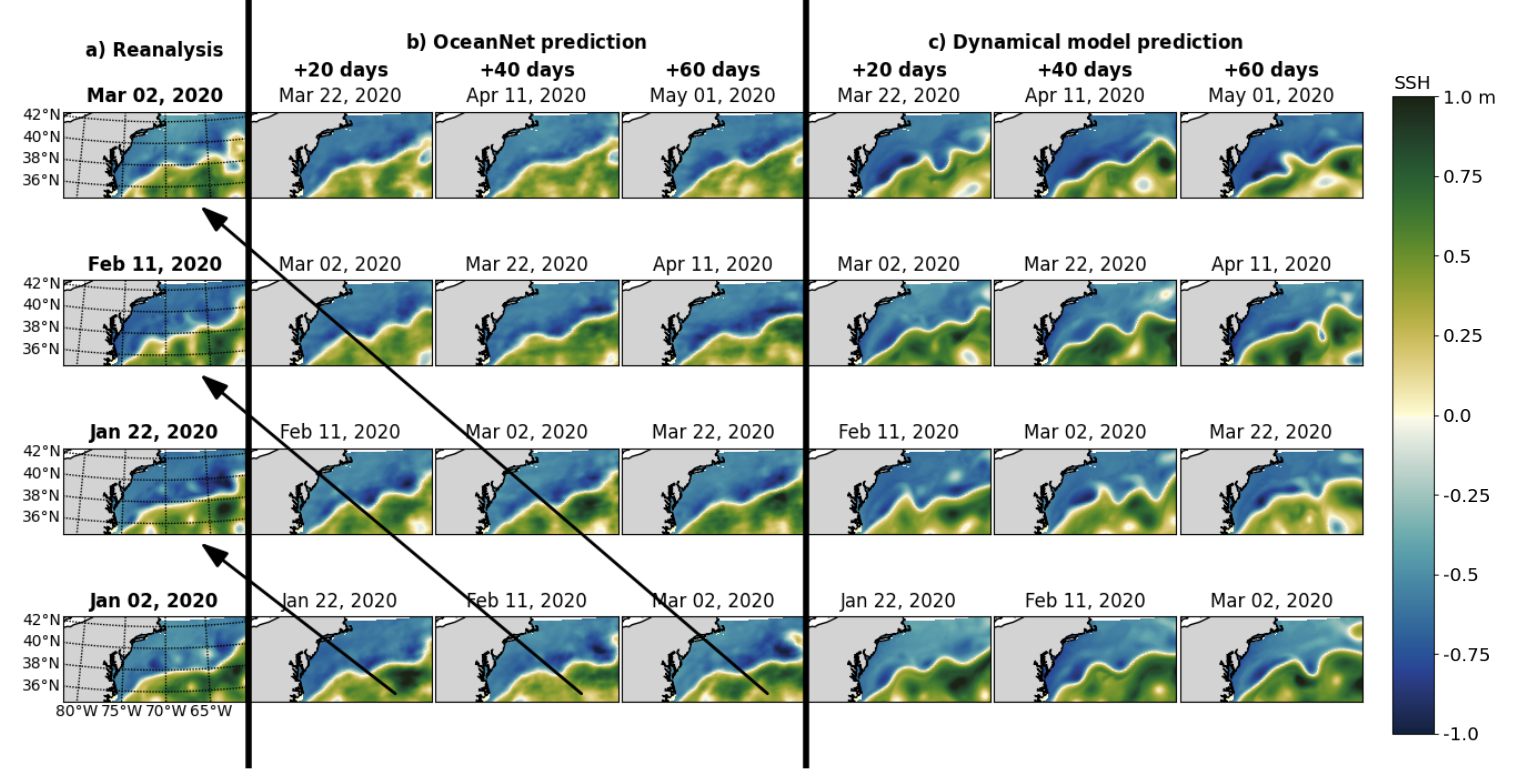}
\caption{Performance of OceanNet for GS prediction. (a) SSH fields from the ocean reanalysis. (b) Predicted SSH generated by OceanNet. (c) Regional ocean dynamical model forecasts. In both OceanNet and dynamical model predictions, each row was initialized with the corresponding reanalysis data in the left column. SSH forecasts are provided for 20, 40, and 60 days. To evaluate the predictions, we can perform a diagonal comparison with the reanalysis SSH, as indicated by the black arrows in (b). The same diagonal comparison can also be conducted with the ocean reanalysis data for (c).}
\label{fig:fig_diagGSM}
\end{sidewaysfigure}

A qualitative assessment reveals that OceanNet effectively captures the SSH propagation of undulations in the northern boundary of the GS. \TT{However, in certain instances it fails to skillfully captures large-scale eddies and traveling into and out of the domain; this can be attributed to the absence of lateral boundary conditions.}~(Fig.~\ref{fig:fig_diagGSM}). \TT{The predictions of OceanNet are also signficantly more diffusive than reanalysis which suggests that OceanNet cannot with just a single variable capture the correct climatology and variability for decadal simulations.} The dynamical model forecast also tends to overpredict extreme values of SSH and the meridional amplitude of the northern boundary. While it is sensitive to initial conditions, OceanNet remains physically consistent over $120$-day forecasts in this region as well. As observed in the GoM, OceanNet provides stable and physically reasonable SSH predictions for the GSM for at least 120 days (not shown for brevity). 

\begin{figure}
\noindent\includegraphics[width=\textwidth]{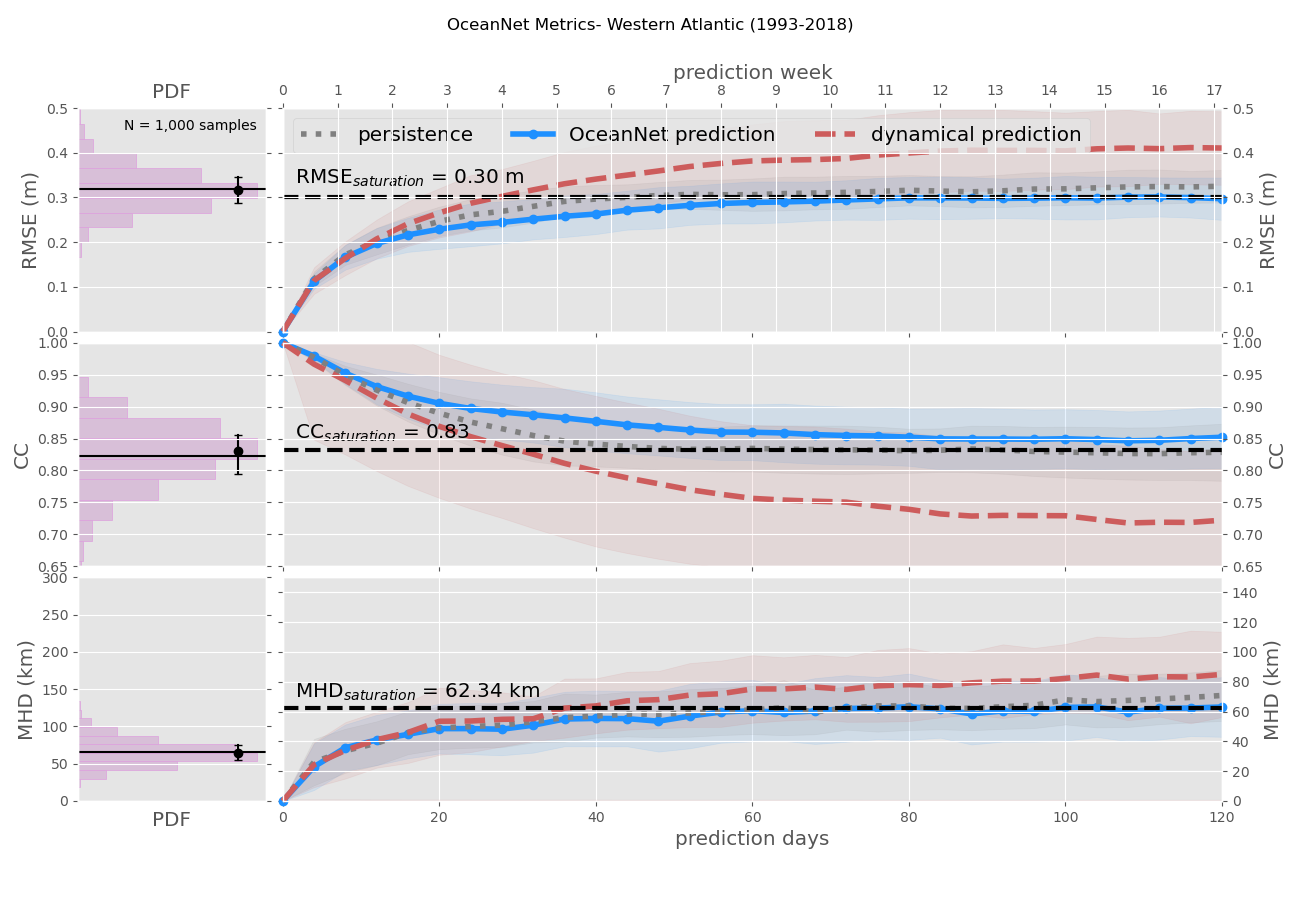}
\caption{OceanNet’s performance metrics in the northwest Atlantic: RMSE (top), CC (middle), and MHD (bottom), compared to the persistence forecast and regional ocean dynamical model forecast. In the left column, probability density functions (PDFs) are presented, derived from 1,000 random pairs sourced from the training data spanning 1993-2018 (means shown by black horizontal lines). The performance statistics, calculated based on forecasts of 0-120 days, are displayed as mean values (lines) with standard deviations (shading). The black horizontal dashed lines denote saturation values, which are determined as 95\% of the means derived from the random pairs. Solid blue lines are OceanNet, dashed red lines are the regional ocean dynamical forecast model, and gray dots are persistence. These representations illustrate how each method’s statistics compare with the target SSH from the reanalysis dataset. \TT{The shading is obtained from the standard deviation due to the choice of different initial conditions to evluate the forecast skill.} }
\label{fig:RMSE_CC_MHD_EC}
\end{figure}

Similar to the GoM, OceanNet consistently outperforms regional ocean dynamical model in RMSE, CC, and MHD computed between the predicted SSH values and the reanalysis SSH over 120 days (Fig. \ref{fig:RMSE_CC_MHD_EC}). Persistence forecasting also fares reasonably well in this region due to the background state of the GSM; however, OceanNet still outperforms persistence in all three metrics over 120 days.

\section{Discussion}\label{sec12}

In this study, we introduce OceanNet, a deep neural operator, with principled design structures within the architecture, to function as a digital twin for long-term (up to 120 days) mesoscale ocean circulation forecasting. Our major technical contributions encompass the development of an FNO and the application of a PEC-based convergent integration scheme, along with a spectral regularizer. Together, these techniques mitigate autoregressive error growth and the small-scale spectral bias frequently observed in data-driven models of multi-scale dynamical systems.

By training the model on 26 years (1993-2018) of high-resolution regional ocean reanalysis data for the northwest Atlantic, we evaluate OceanNet’s ability to predict mesoscale ocean processes such as the GoM LC eddy shedding and GSM over a $90$-$120$ day forecasting horizon. \TT{One major shortcoming of OceanNet is the absence of lateral boundary conditions from the open ocean.} However, since OceanNet is trained using more than two decades of ocean reanalysis data, which considers open ocean processes and conditions, including Rossby waves and ocean eddies propagating into the regional domain, information about ocean boundaries may already embedded in the training dataset. Furthermore, we focus on sub-seasonal to seasonal time scales (30-120 days) in the mid-latitude Atlantic Ocean. The propagation speed of Rossby waves at 40°N is approximately 1 cm/s. As a result, meanders and eddies within our time scale of interest only travel 30-120 km from the open boundary. With this limited travel distance, open ocean condition has a relative small impact on the interior circulation predictions made by the machine learning model.  We acknowledge that for longer-term forecasting problems or regions in lower latitudes (e.g., Rossby wave speed is ~25 cm/s at 5°N), having accurate open boundary conditions will be crucial for model predictions, whether it's a dynamical model or a machine learning model.

Nevertheless, OceanNet demonstrates remarkable stability and consistently \TT{competes with} the regional ocean dynamical model forecast while being $500,000$ times faster. These results demonstrate the potential of utilizing scientific machine learning to develop long-term stable, and accurate data-driven ocean models of great computational efficiency, paving the way for realizing a data-driven digital twin encompassing the entire climate system.

Despite the promising forecast accuracy observed with reanalysis data, our algorithm possesses several limitations. First, this study exclusively trained and tested OceanNet using reanalysis data pertaining to mesoscale ocean eddies and meanders. Real-world ocean forecasting systems operate with dynamical ocean models and real-time ocean observational data, covering dynamic processes across diverse spatial scales. The disparities between these data sources and scales necessitate further investigation into OceanNet’s performance across various ocean applications. Second, OceanNet’s focus is solely on SSH. This study did not explore other ocean state variables, such as currents and temperature. The omission of these variables may restrict the model’s capacity to accurately predict small-scale ocean circulation phenomena, such as shelf break jets and frontal currents. Third, AI-based methods frequently produce smoother forecast results than dynamical models, which can potentially lead to an underestimation of the magnitudes of extreme ocean events. Therefore, additional research is imperative to assess OceanNet’s performance under extreme ocean conditions, e.g., during severe storms. \TT{Finally, while control simulations with OceanNet has been performed in this study, the behavior of OceanNet to perturbations or forcings has not been considered. A physically-consistent AI-based ocean model, akin to physics-based numerical models must produce accurate response to forcings. In future studies, with several variables incorporated into OceanNet, its response to external forcings would also be compared to numerical models to establish physical consistency. }

Significant opportunities exist for improvement in both AI-based methods and dynamical model-based ocean forecasting. In the AI domain, potential advancements involve the integration of subsurface ocean states and additional ocean variables, the incorporation of temporal dimensions through the training of four-dimensional deep networks, and the exploration of more complex network architectures with increased depth and breadth. In the realm of numerical ocean forecast modeling, the development of pre- and post-processing techniques can help mitigate the inherent biases found in ocean models. We expect that a hybrid approach, combining data-driven and dynamical numerical models will play a pivotal role in pushing the boundaries of excellence in ocean prediction.

\section{Data and Methods}\label{sec:data}
\subsection{Reanalysis Data}
\label{sec:ocean_reanalysis}
We utilized high-resolution northwest Atlantic regional ocean reanalysis data to train OceanNet. This reanalysis was generated using a regional implementation of ROMS with the ensemble Kalman filter data assimilation method (EnKF). It features a horizontal resolution of 4 km with 50 vertical layers. For surface atmospheric forcing, we employed data from the European Center for Medium-Range Weather Forecasting (ECMWF) Reanalysis v5 (ERA5), while open boundary conditions were derived from the Copernicus Global Ocean Physics Reanalysis (GLORYS). Ten major tidal constituents from the OSU TPXO tide database were used. The model incorporated 120 river inputs, sourced from the National Water Model and climatology datasets. The temporal scope of the reanalysis spans from January 1, 1993 to December 31, 2020, with daily output.

The assimilated observations encompass a variety of sources, including AVHRR and MODIS Terra sea surface temperature, AVISO along-track sea surface height anomaly, glider temperature/salinity observations from the Integrated Ocean Observing System (IOOS), and the EN4 dataset which aggregates data from Argo floats, shipboard surveys, drifters, moorings, and other sources. EnKF assimilates data with no influence or retainment of previous timesteps, as opposed to 3D/4D-var methods, which include forward and backward passes over the results to ensure continuity. The resulting data assimilative ocean reanalysis allows us to train OceanNet with a time-space continuous SSH dataset with no external knowledge (as would be the case for observations).

OceanNet is trained on 5-day mean SSH data with a 5-day lead time for the GoM domain and 4-day lead time for the GSM domain.  Both training datasets consist of 26 years of reanalysis data from 1993 through 2018. SSH Forecasting was conducted for 2019-2020. Further details on OceanNet’s training process are provided in section~\ref{sec:OceanNet}.

\subsection{OceanNet: Configuration and Design}
\label{sec:OceanNet}
\begin{figure}
\noindent\includegraphics[width=\textwidth]{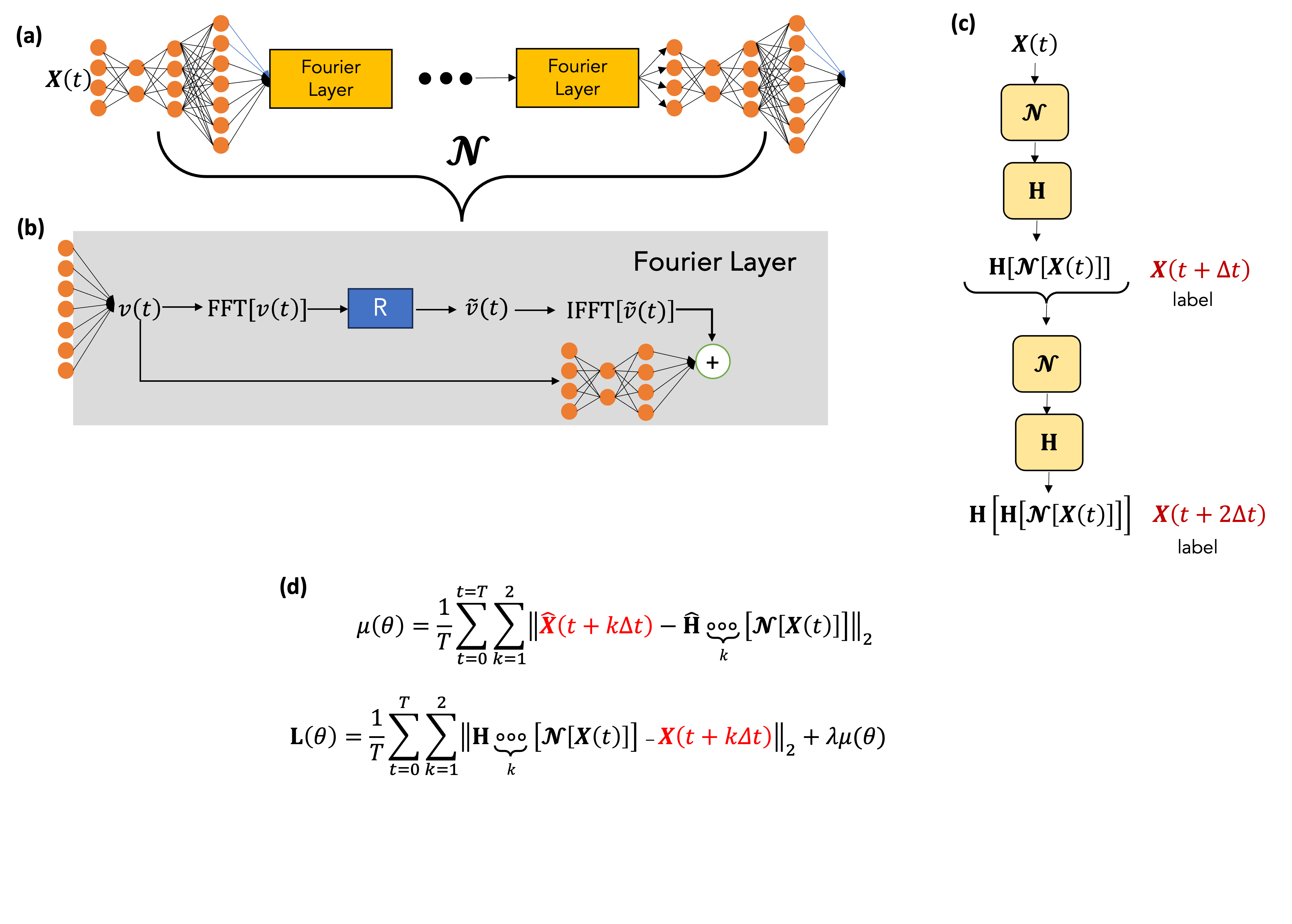}
\caption{(a) A schematic of the OceanNet model. (b) The Fourier Neural Operator, depicted as $\mathcal{N}$. \TT{The neural operator uplifts the input state, $X(t)$ to a high-dimensional space using two convolutional layers. The uplifted state then undergoes a Fourier transform, subsequent coarse-graining by removing the high-wavenumber modes, and then undergoes an inverse Fourier transform. Finally a bias layer is also added to account for aperiodicity in the data. Finally, two more convolutional layers are added to preserve the dimension of the final output, $X(t+k\Delta t)$.} (c) 2-time-step training scheme. (d) The loss function used. Here, $X$ is the state of the system that is predicted, and $\mathbf{H}$ is the PEC-based convergent integration scheme used.}
\label{fig:framework}
\end{figure}

The OceanNet model (Fig.~\ref{fig:framework}) has three major components. the Fourier Neural Operator (FNO), which is used to learn the residual, similar to Krishnapriyan \textit{et al.}~\cite{krishnapriyan2022learning} and Chattopadhyay \textit{et al.}~\cite{chattopadhyay2023long}; the PEC integrator; and the spectral regularizer, which alleviates spectral bias.

\subsubsection{Fourier Neural Operator (FNO)}
\label{sec:FNO}
OceanNet is built upon the FNO~\cite{li2020fourier}. Training utilizes labeled pairs of historical 5-day mean SSH data in the GoM (section~\ref{sec:ocean_reanalysis}), $X(t)$ (input), $X(t+5\Delta t)$ (label), and $X(t+10\Delta t)$ (label), where $\Delta t = 1$ day. For training OceanNet in the northwest Atlantic region, the labels are $X(t+4\Delta t)$ and $X(t+8\Delta t)$. The training assumes the governing partial differential equation for the reduced ocean system involves ocean states $X(t)$:

\begin{align}
\frac{d\mathbf{X}}{dt}= \mathbf{F}\left(\mathbf{X}\left(t\right)\right).
\label{eq:dyn_sys}
\end{align}

To integrate the system from the initial condition, $X(t)$, we represent Eq.~(\ref{eq:dyn_sys}) in its discrete form:
\begin{eqnarray}
     \mathbf{X}(t+5\Delta t)=\underbrace{\mathbf{X}(t)+ 
      \int_{t}^{t+5 \Delta t} \underbrace{\mathbf{F}\left(\mathbf{X}\left(t\right)\right) dt}_{\mathcal{N[\circ,\theta]}}}_{\mathbf{H[\circ]}}. 
      \label{eq:PEC_pde}
\end{eqnarray}
Here, $\mathcal{N}$ is an FNO that parameterizes $\mathbf{F}$ with four Fourier layers, similar to Li et al.~\cite{li2020fourier}, each layer retaining $64$ Fourier modes. $\theta$ represents the $\approx 80\times 10^6$ the trainable parameters of the FNO. $\mathbf{H}$ is a higher-order predictor--evaluate--corrector (PEC) integrator (section~\ref{sec:PEC}). 

\subsubsection{Predictor-Evaluate-Corrector Integration Scheme}
\label{sec:PEC}
Similar to the higher-order integration scheme in the form of fourth-order Runge-Kutta (RK4) (Chattopadhyay et al.~\cite{chattopadhyay2023long}), we implement a PEC scheme in OceanNet, represented by the operator, $\mathbf{H}$. The operations in $\mathbf{H}$ are given by:
\begin{subequations}
\begin{align}
        i_1 &= \mathcal{N}\left[\mathbf{X}\left(t\right),\theta \right], \\
        z &= \mathbf{X}\left(t\right)+\mathcal{N}\left[\mathbf{X}\left(t\right)+\frac{1}{2}i_1,\theta \right]. 
    \end{align}
\end{subequations}
The predicted state is given by $z= \mathbf{H}\left[\mathcal{N}\left(\mathbf{X}\left(t\right)\right),\theta \right]$.

Although most of the higher-order integration schemes, including RK4, demonstrate good performance for this problem, we have identified PEC as the most effective choice. A theoretical study of the effect of each integration scheme on the inductive bias of the trained $\mathcal{N}$ is an active area of research~\cite{krishnapriyan2022learning}, especially for understanding the role it plays on the subsequent spectral bias (section~\ref{sec:fourier_loss}).

\subsubsection{Fourier Regularizer and 2-time-step Loss Function}
\label{sec:fourier_loss}
In OceanNet's loss function, we incorporate a spectral regularizer based on Fourier transforms introduced in Chattopadhyay et al.~\cite{chattopadhyay2023long}. This is in addition to the standard mean squared error loss (MSE) function, which is computed exclusively for grid points located over the ocean. 

The spectral regularizer penalizes deviations in the absolute value of the latitude-averaged zonal Fourier spectrum (Fourier spectrum from now on) of the SSH field at small wavenumbers. Such deviations arise due to spectral bias~\cite{xu2019frequency,chattopadhyay2023long}, which represents an inherent inductive bias in deep neural networks. This bias is responsible for their limitations in learning the fine-scale dynamics of turbulent flow. 

Furthermore, the combined loss function, which integrates both MSE and the spectral regularizer is evaluated over two time steps, namely $t+5 \Delta t$ ($t+4 \Delta t$ for GS) and $t +10 \Delta t$ ($t+8 \Delta t$ for GS), to extend the stability horizon. This two-time-step loss function had remarkable success in the weather prediction model, FourCastNet~\cite{pathak2022fourcastnet}. The spectral regularizers for time $t+5\Delta t$ and $t+ 10\Delta t$ are defined as follows:
\begin{align}
  \mu \left(t+5\Delta t, \theta\right)=\sum_{t=0}^{t=T}\biggr\| \widehat{\mathbf{X}}\left(t+ 5\Delta t\right)\biggr \rvert_{{k_x \geq k_T}}-\widehat{\mathbf{H}}\left[\mathcal{N}\left(\mathbf{X}\left(t\right),\theta\right) \right]\biggr \rvert_{{k_x \geq k_T}}\biggr\|_2^2.
\end{align}

\begin{align}
    \mu \left(t+10\Delta t, \theta\right)=\sum_{t=0}^{t=T}\biggr\| \widehat{\mathbf{X}}\left(t+ 10\Delta t\right)\biggr \rvert_{{k_x \geq k_T}}-\widehat{\mathbf{H}}\left[\mathbf{H}\left[\mathcal{N}\left(\mathbf{X}\left(t\right),\theta\right) \right]\right]\biggr \rvert_{{k_x \geq k_T}}\biggr\|_2^2.
\end{align}

Here, the $\hat{\left[\circ\right]}$ operator is the latitude-averaged zonal Fourier transform of both the predicted and the target SSH fields. $k_T$ is a cutoff wavenumber chosen to be $100$ after significant trial and error. 
The  loss functions for $t+5 \Delta t$ and $t+10\Delta t$ are given by $\mathbf{L}\left(t+5\Delta t, \theta\right)$:

\begin{align}
\label{eq:FourK_spectal_loss_1}
    \mathbf{L}\left(t+5 \Delta t, \theta\right)=\sum_{t=0}^{t=T}\|\mathbf{X}\left(t+5\Delta t\right)- \mathbf{H}\left[\mathcal{N}\left(\mathbf{X}\left(t\right),\theta\right) \right]\|_2^2 + \lambda \mu(t+5\Delta t,\theta).
\end{align}

\begin{align}
\label{eq:FourK_spectal_loss_2}
\mathbf{L}\left(t+10 \Delta t, \theta\right)=\sum_{t=0}^{t=T}\|\mathbf{X}\left(t+10\Delta t\right)- \mathbf{H}\left[\mathbf{H}\left[\mathcal{N}\left(\mathbf{X}\left(t\right),\theta\right) \right]\right]\|_2^2 + \lambda \mu(t+10\Delta t,\theta).
\end{align}

The total loss is given by:
\begin{align}
\label{eq:total_loss}
    \mathbf{L}\left(\theta\right)=\mathbf{L}\left(t+5 \Delta t, \theta\right)+\mathbf{L}\left(t+10 \Delta t, \theta\right).
\end{align}

The spectral regularizer ensures that the high wavenumbers in the Fourier spectrum of SSH remain consistent with the target Fourier spectrum. This measure is effective in reducing spectral bias throughout the training process. The two-time-step loss function enhances the stability horizon of the model. The number of time steps over which the loss is calculated can potentially be extended. However, with each increase in the number of time steps, the memory requirement for the subsequent backpropagation process during training grows exponentially.


\bmhead{Acknowledgments}
We thank Subhashis Hazarika and Maria Molina for the insightful discussions and Jennifer Warrillow for editorial assistance. MG, AL, TW, and RH, were supported by NSF awards 2019758 and 2331908. AC performed partial work at the Palo Alto Research Center, SRI. All authors contributed equally to the research conducted and the writing of the paper. The codes used in this study are openly available at \url{https://github.com/magray-ncsu/OceanNet}.


\bibliography{sn-bibliography}


\end{document}